\documentclass[a4paper, 10pt, conference]{ieeeconf}      
\usepackage{FG2024}

\FGfinalcopy 

\IEEEoverridecommandlockouts                              
\usepackage{graphicx}
\usepackage{amsmath}
\usepackage{url} 
\usepackage{hyperref}
\usepackage{authblk}

\def\FGPaperID{6} 

\title{\LARGE \bf
Analyzing the Effect of Combined Degradations on Face Recognition
}

\usepackage{fancyhdr}
\begin{document}

\ifFGfinal
\thispagestyle{empty}
\author[1]{Erdi Sarıtaş}
\author[1,2]{Hazım Kemal Ekenel}
\affil[1]{\small Department of Computer Engineering, Istanbul Technical University, Istanbul, Türkiye}
\affil[2]{\small Department of Computer Science and Engineering, Qatar University, 
Doha, Qatar}
\pagestyle{empty}
\else
\author{Anonymous FG2024 submission\\ Paper ID \FGPaperID \\}
\pagestyle{plain}
\fi
\maketitle

\thispagestyle{fancy}

\begin{abstract}

A face recognition model is typically trained on large datasets of images that may be collected from controlled environments. This results in performance discrepancies when applied to real-world scenarios due to the domain gap between clean and in-the-wild images. Therefore, some researchers have investigated the robustness of these models by analyzing synthetic degradations. Yet, existing studies have mostly focused on single degradation factors, which may not fully capture the complexity of real-world degradations. This work addresses this problem by analyzing the impact of both single and combined degradations using a real-world degradation pipeline extended with under/over-exposure conditions. We use the LFW dataset for our experiments and assess the model's performance based on verification accuracy. Results reveal that single and combined degradations show dissimilar model behavior. The combined effect of degradation significantly lowers performance even if its single effect is negligible. This work emphasizes the importance of accounting for real-world complexity to assess the robustness of face recognition models in real-world settings. The code is publicly available at \href{https://github.com/ThEnded32/AnalyzingCombinedDegradations}{https://github.com/ThEnded32/AnalyzingCombinedDegradations}.
\end{abstract}

\section{INTRODUCTION}

Facial recognition systems have evolved significantly, due to advancements in deep learning research and available large-scale datasets \cite{webface, ms1m}. due to the ease of access to advanced deep learning resources and large and diverse datasets, the researchers have created better models on the benchmarks \cite{fr_survey}. In addition to enabling model comparisons, benchmarks also help to identify open problems.

İnitial face recognition datasets and benchmarks were designed on simpler and more controlled where face images are clean and professionally taken \cite{feret, arface, yale}. As face recognition models have achieved promising results on these benchmarks, researchers have begun exploring their applications in more difficult and uncontrolled environments. This results in the necessity for more challenging benchmarks, such as real-world (in-the-wild) benchmarks that aim to capture the complexities and variations of real-world situations \cite{lfw,ijbC}. Then, the researchers focused on improving performance regarding real-world benchmarks by collecting wilder datasets and/or building more robust systems.

Some researchers have analyzed how the models behave if the inputs are degraded synthetically \cite{deg_analysis1, deg_analysis2}. These studies have provided insights to distinguish the effects of the different types of degradation. Nonetheless, real-world degradations are combinations of different types of degradations occurring simultaneously. Therefore, investigating each degradation in isolation may discard the relation and fused effect of combined degradations. Consequently, as each degradation was tested individually, their approach may fail to capture the mutual effects that occur in real-world scenarios. Therefore, analyzing more complex degradations is important, since relying on single degradations may lead to incorrect assumptions to verify model robustness.

This study examines the effect of combined degradations as well as single ones on face recognition performance. We mimic real-world degradations with a commonly used degradation pipeline that combines multiple synthetic degradations. Moreover, exposure changes are integrated into this pipeline. Our primary objective is to assess face recognition performance under challenging conditions, where multiple degradations occur together, such as in surveillance. We aim to provide an in-depth understanding of the effect of those combined degradations. Furthermore, our analysis highlights differences in a face recognition model's behavior when encountered with single and combined degradations.

\section{Related Works}

Deep learning has appeared as a key milestone in performance advancements for face recognition models. These models have been trained on very large datasets using specialized loss functions to extract discriminative features to represent faces, where these features are utilized for verification or identification tasks \cite{fr_survey}. 

Since face recognition has become more prevalent in daily life, the robustness of these models is essential. Earlier, face recognition models were trained and evaluated on datasets collected from controlled environments \cite{feret, arface, yale}. Despite significant progress, the discrepancy between real-world scenarios and the used training data causes a domain gap, which can reduce the models' performance. To minimize the impact of this gap and improve representation learning, some works constructed in-the-wild datasets \cite{webface} and benchmarks~\cite{lfw, ijbC} that include more diversity in terms of ethnicity, gender, age, head poses, occlusions, and low-quality images. 

These datasets have helped models to generalize better and extract more discriminate features. However, the domain gap problem could not be resolved completely despite collecting images under uncontrolled environments. Recent studies have addressed that unseen data biases can impact performance \cite{fr_survey}. Several studies have examined biases, including demographic factors \cite{bias} like ethnicity, gender, and age. Some researchers have focused on image quality aspects such as occlusion \cite{occluded} or degradation \cite{deg_analysis1, deg_analysis2, deg_analysis_gibi, deg_analizWcombine,degWvit}.

Imitating real-world degradations synthetically is a common strategy that has been employed for different tasks ranging from cross-resolution face recognition \cite{crossFRWdeg, meetinthemiddle} to super-resolution \cite{realesrgan}. Researchers have applied this approach to gain insights into how real-world degradations impact model performance \cite{deg_analysis1, deg_analysis2, deg_analysis_gibi, deg_analizWcombine,degWvit}. Synthetic degradation alternatives like convolving with a blur filter or adding Gaussian noise are applied to analyze how degradations like motion blur or noise affect the models' performance.

Studies like \cite{deg_analysis1, deg_analysis2} followed the synthetic degradation technique to examine the influence of various degradations on different face recognition models. They also assessed the impact of missing data by synthetically masking vital facial regions, such as eyes, mouth, etc. Although these studies offer valuable insights, they may miss real-world complexity as the combined degradations are commonly encountered. \cite{deg_analizWcombine} analyzed combined degradations, yet only tested a limited number of combinations with the same difficulty level. 

\section{METHODOLOGY}

This work analyzes the effect of degradations encountered in real-world scenarios. Considering the variations in \cite{realesrgan, meetinthemiddle} such as applying the degradation pipeline multiple times or ordering the degradations differently, the degradation combinations used in previous works~\cite{deg_analizWcombine, deg_analysis1, deg_analysis2} look insufficient. Therefore, we focused on extending the combination space as much as possible. Although the real-world degradations are much more complex, we approximated it with the widely used degradation pipeline \cite{realesrgan} as,

\begin{equation}
    I_{LR} = [(I_{HR} \otimes k_{blur})\downarrow_r + N_{additive}]_{JPEG}
\label{eq-degredation}
\end{equation}

\noindent where $I_{HR}$ is the high-resolution image. The $I_{HR}$ is first convolved with a blur kernel $k_{blur}$, then downsampled by the ratio $r$. Later, the additive noise $N_{additive}$ is added, and lastly, JPEG compression artifacts are inserted to create a synthetically degraded low-resolution image $I_{LR}$. 

Since under/over-exposure commonly occurs in real-world situations, we include abnormal exposure settings in the pipeline by applying the equation from \cite{exposure_formül}:

\begin{equation}
    I_{Exposured} = \begin{cases}
    (1-(1-I_{HR}/255)^\gamma)*255, \text{ if 0-255 range}\\
    1-(1-I_{HR})^\gamma, \text{ if 0-1 range} 
\end{cases}
\label{eq-exposure}
\end{equation}

\begin{equation}
    I_{Exposured-LR} = [(I_{Exposured} \otimes k_{blur})\downarrow + N_{additive}]_{JPEG}
\label{eq-degredationWexposure}
\end{equation}

\begin{figure}[!t]
	\centering
    \includegraphics[width=0.75\linewidth]{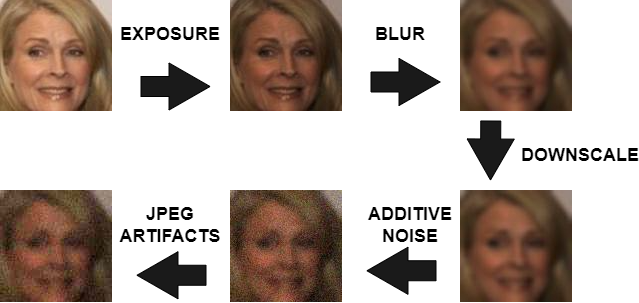}
    \caption{Illustration of degradation pipeline. First, under/over-exposure is applied. Then, a (Gaussian) kernel is used for blurring. Following, (bicubic) downscaling is used and additive (Gaussian) noise is added. Lastly, JPEG artifacts are inserted.}
	\label{fig-pipeline}
\end{figure}

 As the exposure is more related to the environment itself rather than the camera or the objects of interest, under/over-exposure is applied to the $I_{LR}$ with the gamma value $\gamma$ in (\ref{eq-exposure}) beforehand. Then, $I_{Exposured}$ is fed to the mentioned degradation pipeline to compose $I_{Exposured-LR}$ in (\ref{eq-degredationWexposure}). The visualization of the overall pipeline, including the exposure, can be seen in Fig.~\ref{fig-pipeline}. Even though the person can be identified easily at the first one or two degradation steps, the end result is much more challenging to recognize.

\section{EXPERIMENTS}

In the experiments, we used the ArcFace \cite{arcface} with ResNet100 \cite{resnet} backbone trained on MS1MV3 \cite{ms1m} as the face recognition model, and LFW~\cite{lfw} as the testing dataset. In the evaluations, we followed the 10-fold cross-validation procedure of the LFW. The baseline verification run with the non-degraded (original) images is done for threshold optimization. The specified threshold values are recorded and used in the later verification runs. Therefore, no threshold optimization is made in the runs with the degraded images.

Table~\ref{tab-degparams} presents the used degradation parameters. Every possible degradation combination is applied using all the parameters. We employed two verification versions, 'normal' and 'cross'. In the 'normal' verification, both images of the pairs to be compared are degraded. While in the 'cross' verification, one of the compared images is kept as the original (without degrading). After arranging the images, we extracted their features and performed the verification using the recorded thresholds. The average verification result of the 10-fold cross-validation is used for the model performance. Moreover, since the noise addition was non-deterministic, we repeated the same experiment -- five times if $\sigma_{Noise}$ is not $None$ -- and averaged their results. 

\begin{table}[!b]
\caption{Degradation Parameters}
\begin{center}
    \resizebox{0.85\linewidth}{!}{
    \begin{tabular}{|c||c|}
        \hline
         \textbf{Degradation Type}& \textbf{Parameters} \\ \hline
         $\sigma_{Noise}$ & None,2,4,8,16,32,\textbf{64} \\ \hline
         $JPEG-Quality$ & None,64,32,16,8,\textbf{4} \\ \hline
         $Downscale Ratio (r)$ & None,2,3,4,\textbf{8} \\ \hline
         $Kernel Parameters$ & None,(1,1,0),(2,2,0),\textbf{(3,3,0)}, \\ 
         $(\sigma_x,\sigma_y,\theta_{rotation})$ & (1,3,0), (1,3,$\pi$/4),(1,3,$\pi$/2),(1,3,3$\pi$/4) \\ \hline
         $\gamma_{Exposure}$ & \textbf{0.125},0.25,0.5,None,2,4,\textbf{8} \\ \hline
     \multicolumn{2}{l}{$^{\mathrm{*}}$ Extreme cases are marked with \textbf{bold}.}
    \end{tabular}}
\label{tab-degparams}
\end{center}
\end{table}

\begin{figure}[!b]
    \centering
    \resizebox{\linewidth}{!}{
    \begin{tabular}{ccc}
        (1,1,0) &  (2,2,0) & (3,3,0)\\
        \includegraphics[width=0.1\linewidth]{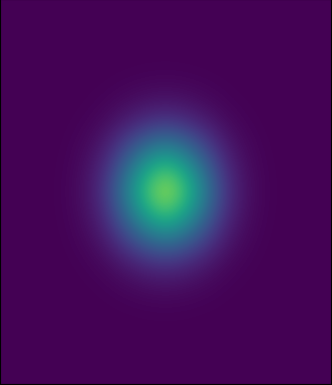} & 
        \includegraphics[width=0.1\linewidth]{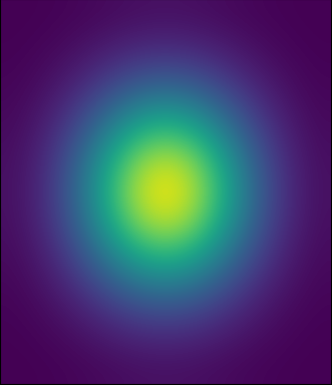} & 
        \includegraphics[width=0.1\linewidth]{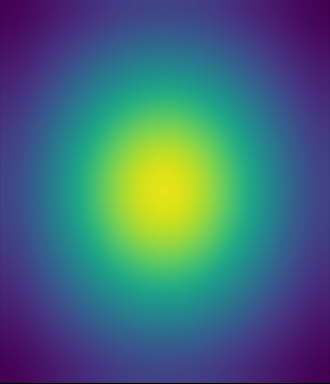} 
         \\
    \end{tabular}
    \begin{tabular}{cccc}
        (1,3,0) & (1,3,$\pi$/4) & (1,3,$\pi$/2) &(1,3,3$\pi$/4)\\
        \includegraphics[width=0.1\linewidth]{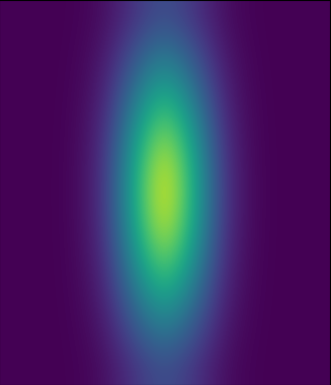} & 
        \includegraphics[width=0.1\linewidth]{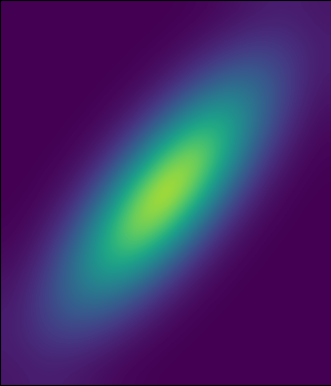} & 
        \includegraphics[width=0.1\linewidth]{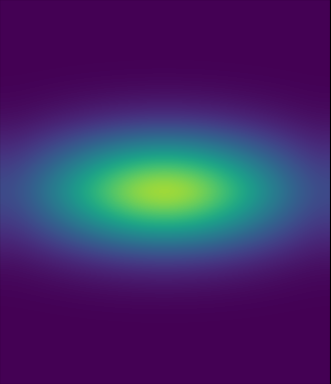} & 
        \includegraphics[width=0.1\linewidth]{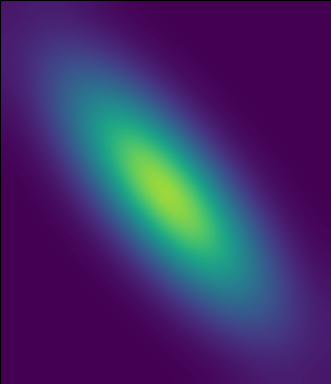} 
        \\
    \end{tabular}}
    \caption{The $k_{blur}$ kernels with their parameters on top.}
    \label{fig:blur-kernels}
\end{figure}

 We built the degradation pipeline using the BasicSR \cite{basicsr} repository. For $k_{blur}$, an anisotropic bivariate Gaussian kernel with $11\times11$ window size is used, and the isotropic flag is set to false. Table~\ref{tab-degparams} shows Gaussian kernel parameters $\sigma_x$ and $\sigma_y$, and rotation parameter $\theta_{rotation}$. Fig. \ref{fig:blur-kernels} shows the applied kernels. For the additive noise, Gaussian noise is sampled from the Normal distribution with the size of the input and then scaled with the $\sigma$ value. Downscale operation is done using the bicubic filter. The DiffJPEG is used for the JPEG artifacts. After completing the degradation pipeline, the images are resized to a resolution of $112\times112$ pixels.

\section{RESULTS}

\begin{figure}[!b]
	\centering
{\includegraphics[width=0.78\linewidth]{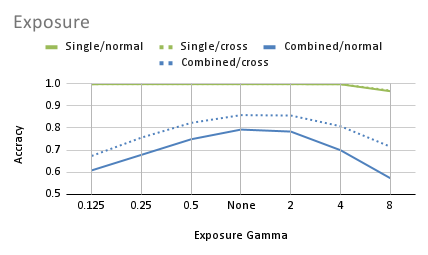} } \\
{\includegraphics[height=55pt]{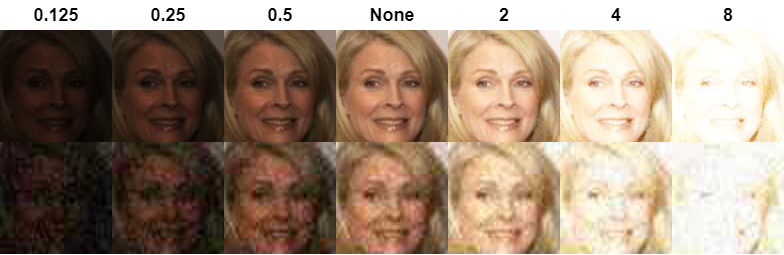} }  \\
    \caption{Exposure setting results. Examples of single (top) and combined (bottom) degradation results are shown in the below sub-figure.}
	\label{fig-exposure}
\end{figure}

\begin{figure}[!b]
	\centering
{\includegraphics[width=0.78\linewidth]{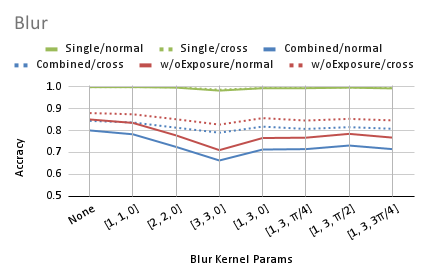} } \\
{\includegraphics[height=57pt]{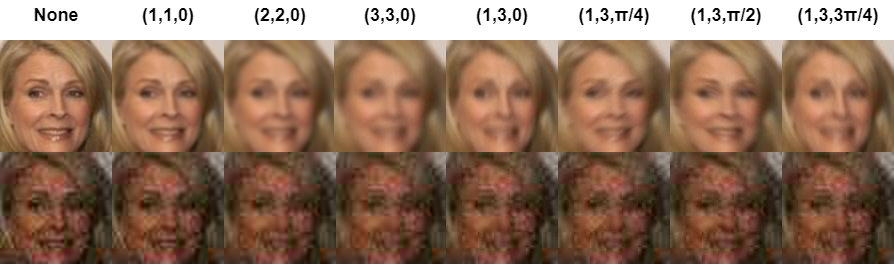} }  \\
    \caption{Blurring results. Examples of single (top) and combined (bottom) degradation results are shown in the below sub-figure.}
	\label{fig-blur}
\end{figure}

\begin{figure}[!t]
	\centering
{\includegraphics[width=0.78\linewidth]{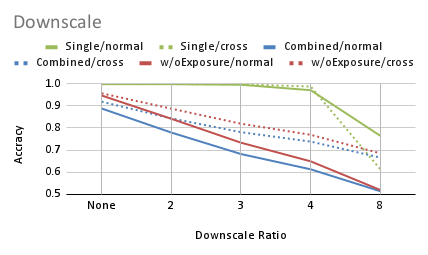} } \\
{\includegraphics[height=57pt]{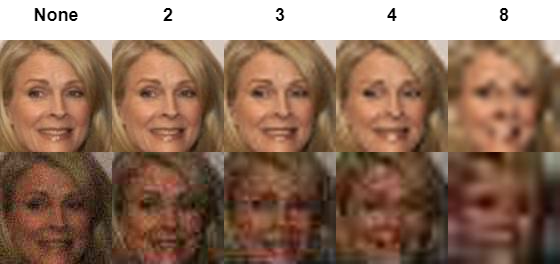} }  \\
    \caption{Downscaling results. Examples of single (top) and combined (bottom) degradation results are shown in the below sub-figure.}
	\label{fig-downscale}
\end{figure}

\begin{figure}[!b]
	\centering
{\includegraphics[width=0.78\linewidth]{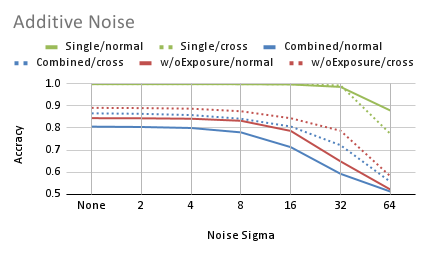} } \\
{\includegraphics[height=57pt]{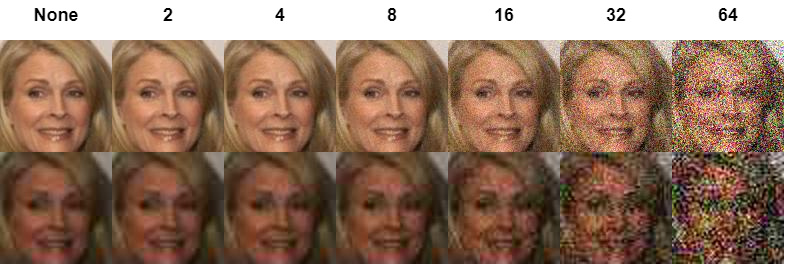} }  \\
    \caption{Additive noise results. Examples of single (top) and combined (bottom) degradation results are shown in the below sub-figure.}
	\label{fig-noise}
\end{figure}

\begin{figure}[!t]
	\centering
{\includegraphics[width=0.78\linewidth]{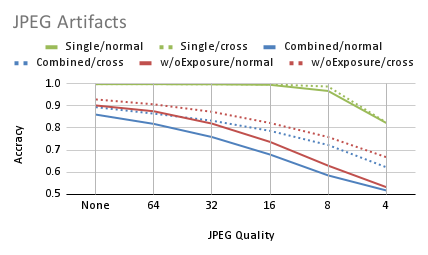} } \\
{\includegraphics[height=57pt]{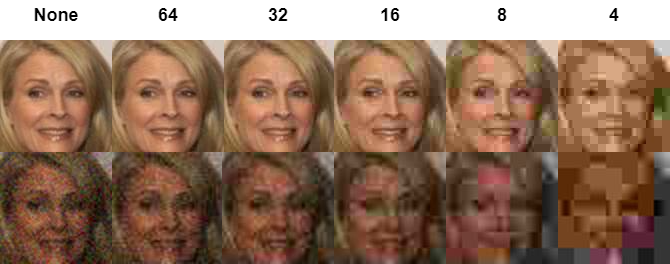} }  \\
    \caption{JPEG artifact insertion results. Examples of single (top) and combined (bottom) degradation results are shown in the below sub-figure.}
	\label{fig-jpeg}
\end{figure}

Fig. \ref{fig-exposure}-\ref{fig-jpeg} presents experimental results. The results show the effects of both 'single' and 'combined' degradation on verification accuracy. Only the chosen degradation is applied with its selected parameters for the 'single' degradation results. The averages of the results from all combinations for the chosen degradation are shown for the 'combined' degradation results. To show the results with the degradation pipeline in \cite{realesrgan}, results without under/over-exposure were presented as 'w/oExposure' explicitly in Fig.~\ref{fig-blur}-\ref{fig-jpeg}. In each figure, the top sub-figure shows the verification accuracy graph, and the bottom sub-figure shows example results from applied 'single' (top) and 'combined' (bottom) degradations to a face image sample.

The combined and w/oExposure accuracies were found to be much lower than our expectations during the preliminary analysis. We found that when multiple extreme cases -- marked as \textbf{bold} in Table \ref{tab-degparams} -- are encountered together, the accuracies become close to 0.5. Moreover, we observed unrecognizable images when examining sample outputs from these extreme parameter combinations. Therefore, to eliminate unrealistic or extremely rare scenarios from evaluation, we omitted results from degradations whose parameters include multiple extreme cases while aggregating them for 'combined' and 'w/oExposure' accuracies. With this filtering, we aggregated results with at most one extreme case.

The effect of under/over-exposure is discussed first as it is the first applied degradation. From Fig. \ref{fig-exposure}, it can be seen that applying the exposure individually does not affect the accuracy much except for the gamma value eight where there is a slight decrease. However, under-exposure and over-exposure decrease the accuracy almost symmetrically when considering the effect of combined degradation. Regarding the verification types, single results are almost identical, while cross results are nearly swiped-up versions of the normal ones. If we look at the sample degraded face images in Fig. \ref{fig-exposure}, even with the gamma values 0.25 or 4 the person can be recognized easily for the single degradation. However, for the combined versions, it can be said that most of the details are lost.

Fig.~\ref{fig-blur} displays the effect of blurring. The single degradation has almost no effect where both combined and w/oExposure have a significant accuracy drop especially for the normal verification scenario. When we analyze the results of the normal verification scenario in detail, we observe that there is a significant performance change for the circular kernels proportional to the size of the kernel. Moreover, for the elliptic kernels, vertical and horizontal kernels with parameters (1,3,0) and (1,3,$\pi$/2) decrease the similarity less than the other two. For the cross verification, the results are closer to each other and almost show a flat behavior. Moreover, the accuracies are also higher. w/oExposure results are roughly swiped-up versions of the combined ones. The sample degraded face images in Fig.~\ref{fig-blur} show that faces are mostly recognizable, except the one degraded with (3,3,0) kernel. However, these small differences lead to more important detail loss with the combined degradations.

When it comes to downscaling, the single effect nearly appears only at the downscale ratio of eight in Fig. \ref{fig-downscale}. Different from previous results, there is a significant margin between normal and cross verification results for the downscale ratio of eight. Moreover, both the combined and w/oExposure show a linear-like behavior for both normal and cross verification where accuracy decreases less in the cross verification scenario. Considering the example face images below in Fig.~\ref{fig-downscale}, the single downscale with a ratio of eight damages the facial details. In addition, the main facial features are spoiled with the combined degradation even for the downscale ratios of less than eight.
\pagebreak

Fig. \ref{fig-noise} exhibits the effect of additive noise. The performance almost does not change until the sigma value reaches 64 for single cases, whereas the cross verification obtains lower accuracy than the downscale. For the combined and w/oExposure cases, performance drops fairly after the sigma value of 16. A similar behavior with a small accuracy margin can be seen from the cross verification results. However, the noise sigma of 32 decreases performance more with normal verification. If we look at the examples below in Fig. \ref{fig-noise}, even single degradation with the sigma value of 64 the person's identity can be recognized. However, for the combined version, the face becomes unrecognizable.

For the JPEG artifact, Fig.~\ref{fig-jpeg} shows the results. The accuracy of the single case drops dramatically after the JPEG quality of eight. The combined and w/oExposure results follow a consistent decrease as the JPEG quality decreases for both normal and cross verification scenarios where cross verification has less change. From the example face images in Fig.~\ref{fig-jpeg}, it can be observed that only the JPEG quality of four is hard to identify when applied in isolation, while even the JPEG quality of eight led to loss of facial features for the combined degradation.

\section{CONCLUSIONS AND FUTURE WORKS}

This work conducted a comprehensive analysis of the behavior of a face recognition model under various real-world degradations. Acknowledging the limitations of synthetic degradations on capturing the complexity of real-world conditions, we utilized a commonly used degradation pipeline that combines multiple degradation factors. Furthermore, we extended the pipeline so that abnormal exposure settings were incorporated. The LFW dataset and an ArcFace model were used for the experiments. Both the normal and cross verification accuracies are considered for the analysis. Moreover, we provided visual examples to understand better how the degradations affect the images. 

Our findings highlight that there is a substantial difference in the model's response to single versus combined degradations. Despite the minimal effect of isolated degradation, degradations significantly deteriorate performance when combined. Furthermore, the model shows different behaviors between normal and cross verification scenarios for single and combined degradations. From the results, it can be concluded that combined degradations should also be considered in addition to single ones to assess the robustness of face recognition models.

Further investigations could explore the impact of more complex synthetic degradations like second-order degradations or different ordering. Moreover, different versions of degradations, like Salt\&Pepper noise, could be integrated. In addition to these, similar to previous works, the effect of missing facial areas could be incorporated. Also, the correlation between the degradations should be investigated.

\section{ACKNOWLEDGMENTS}

\noindent \small We would like to thank Dr. Yusuf Hüseyin Şahin and Ali Azmoudeh for their valuable comments. This work was supported by the TUBITAK BIDEB 2210 Graduate Scholarship Program and ITU BAP (Project No: 44915).


\pagebreak

{\small
\bibliographystyle{ieee}
\bibliography{egbib}
}

\end{document}